\documentclass[pdflatex,sn-mathphys-num]{sn-jnl}

\usepackage{booktabs}       
\usepackage{multirow}       
\usepackage{amssymb}        
\usepackage{float}         
\usepackage{amsmath,amsfonts}
\usepackage{algorithm}
\usepackage{array}
\usepackage[caption=false,font=normalsize,labelfont=sf,textfont=sf]{subfig}
\usepackage{textcomp}
\usepackage{stfloats}
\usepackage{url}
\usepackage{verbatim}
\usepackage{graphicx}

\usepackage{hyperref}
\usepackage{verbatim}
\usepackage{graphicx}
\usepackage{booktabs}
\usepackage[table]{xcolor}
\usepackage{booktabs}
\usepackage{multirow}
\usepackage{pifont}
\usepackage{makecell}

\usepackage{capt-of}

\usepackage{graphicx}

\hypersetup{
    colorlinks=true,
    linkcolor=blue,    
    citecolor=blue,      
    urlcolor=blue       
}
\usepackage{graphicx}%
\usepackage{multirow}%
\usepackage{amsmath,amssymb,amsfonts}%
\usepackage{amsthm}%
\usepackage{mathrsfs}%
\usepackage[title]{appendix}%
\usepackage{xcolor}%
\usepackage{textcomp}%
\usepackage{manyfoot}%
\usepackage{booktabs}%
\usepackage{algorithm}%
\usepackage{algorithmicx}%
\usepackage{algpseudocode}%
\usepackage{listings}%
\usepackage{fancyvrb}
\usepackage{xcolor}

\newcommand{\promptbox}[2][white]{%
\medskip
  \begin{minipage}{0.999\linewidth}
    \fcolorbox{black}{#1}{%
      \parbox{\dimexpr0.999\linewidth-2\fboxsep-2\fboxrule}{%
        \fontsize{6}{6}\ttfamily\selectfont

        \hangindent=2em
        #2
      }%
    }
  \end{minipage}%
\medskip
}

\theoremstyle{thmstyleone}%
\theoremstyle{thmstyletwo}%

\theoremstyle{thmstylethree}%

\begin{document}

\title[Article Title]{HyCoVAD: A Hybrid SSL-LLM Model for Complex Video Anomaly Detection}


\author[1]{\fnm{Mohammad Mahdi} \sur{Hemmatyar}}\email{mahdi.hemmat@sharif.edu}
\author*[1]{\fnm{Mahdi} \sur{Jafari Siavoshani}}\email{mjafari@sharif.edu}

\author[2]{\fnm{Mohammad Amin} \sur{Yousefi}}\email{m.amin.yousefi@ut.ac.com}
\equalcont{These authors contributed equally to this work.}

\author[2]{\fnm{Mohammad Reza} \sur{Nemati}}\email{mohammad.r.nemati@ut.ac.ir}
\equalcont{These authors contributed equally to this work.}

\author[3]{\fnm{Mobin} \sur{Azadani}}\email{m\_azadani@comp.iust.ac.ir}
\equalcont{These authors contributed equally to this work.}

\author[3]{\fnm{Hamid Reza} \sur{Rastad}}\email{h\_rastad@elec.ac.iust.ir}
\equalcont{These authors contributed equally to this work.}

\author[3]{\fnm{Amirmohammad} \sur{Akbari}}\email{am\_akbari@elec.iust.ac.ir}
\equalcont{These authors contributed equally to this work.}

\affil*[1]{\orgdiv{Department of Computer}, \orgname{Sharif University of Technology}, \orgaddress{\street{Azadi st.}, \city{Tehran}, \postcode{11155-9517}, \country{Iran}}}

\affil[2]{\orgdiv{School of Electrical and Computer Engineering}, \orgname{University of Tehran}, \orgaddress{\street{North Kargar st.}, \city{Tehran}, \postcode{14395-515}, \, \country{Iran}}}

\affil[3]{\orgdiv{School of Computer Engineering}, \orgname{Iran University of Science and Technology}, \orgaddress{\street{Hengam st.}, \city{Tehran}, \postcode{16846-13114}, \country{Iran}}}

\abstract{
Video anomaly detection (VAD) is crucial for intelligent surveillance, but a significant challenge lies in identifying complex anomalies, which are events defined by intricate relationships and temporal dependencies among multiple entities rather than by isolated actions. While self-supervised learning (SSL) methods effectively model low-level spatiotemporal patterns, they often struggle to grasp the semantic meaning of these interactions. Conversely, large language models (LLMs) offer powerful contextual reasoning but are computationally expensive for frame-by-frame analysis and lack fine-grained spatial localization.
We introduce HyCoVAD, Hybrid Complex Video Anomaly Detection, a hybrid SSL-LLM model that combines a multi-task SSL temporal analyzer with LLM validator. The SSL module is built upon an nnFormer backbone which is a transformer-based model for image segmentation. It is trained with multiple proxy tasks, learns from video frames to identify those suspected of anomaly. The selected frames are then forwarded to the LLM, which enriches the analysis with semantic context by applying structured, rule-based reasoning to validate the presence of anomalies.
Experiments on the challenging ComplexVAD dataset show that HyCoVAD achieves a 72.5\% frame-level AUC, outperforming existing baselines by 12.5\% while reducing LLM computation. We release our interaction anomaly taxonomy, adaptive thresholding protocol, and code to facilitate future research in complex VAD scenarios.}

\keywords{Video anomaly detection, complex anomaly, self-supervised learning, large language models, video language models, embedding model}




\maketitle

\section{Introduction}\label{sec1}
Video anomaly detection (VAD) is a fundamental problem in computer vision with wide-ranging applications in surveillance, safety monitoring, and autonomous systems. The central objective is to identify events that deviate from normative visual patterns. Early research in this domain emphasized anomalies characterized by low-level irregularities such as unusual trajectories or motion patterns, but they primarily captured simple, single-agent deviations.
In real-world environments, however, anomalies are often defined by complex interactions among multiple entities rather than by isolated behaviors. Examples include verbal disputes escalating into physical altercations, unattended packages in crowded spaces, or vehicles violating coordinated traffic flow. These interaction-driven anomalies present unique challenges: they emerge from higher-order relationships between agents, unfold over extended temporal horizons, and require context-sensitive reasoning that bridges perception and semantics. To address these challenges, the ComplexVAD dataset \cite{mumcu2025complexvad} was recently introduced, encompassing 40 interaction anomaly classes across 217 videos. Baselines on this dataset, as shown in \cite{mumcu2025complexvad}, reveal a low performance for state-of-the-art methods (60\% AUC), highlighting a persistent gap between conventional anomaly detection paradigms and the demands of interaction reasoning.

Two complementary research directions have attempted to bridge this gap. Self-supervised learning (SSL) has become a dominant paradigm for VAD, leveraging proxy tasks such as motion prediction, frame reconstruction, and spatial puzzles to learn normal spatiotemporal patterns from unlabeled video. SSL methods are efficient and generalize well to unseen environments. However, recent surveys \cite{Raja2023, liu2024llavanext} emphasize their limitations in hierarchical reasoning: while they can capture atomic visual actions, they struggle to infer semantic meaning in multi-agent interactions.

As a second approach, the advent of large language models (LLMs) has opened new opportunities for vision-language reasoning in video. Training-free approaches leverage LLMs to classify or explain anomalies based on captions or symbolic rules. These methods excel at contextual interpretation but face two critical barriers: (1) high computational cost, as billions of parameters are applied to redundant video frames, and (2) limited spatial precision, as subtle anomalies often evade text-based reasoning. 

This paper proposes to unify the efficiency of SSL with the semantic reasoning of LLMs in a detect-validate hybrid framework. Our approach leverages SSL as a proposal generator for potential anomalies and an LLM as a semantic validator for uncertain cases. 

The contributions of this work are threefold:
\begin{itemize}
    \item We implemented a hybrid anomaly detection framework that integrates an SSL module with a training-free LLM-based stage for complex, interaction-driven events.
    \item We introduced a coarse-to-fine filtering mechanism where an efficient SSL module acts as an anomaly proposal generator, reducing the number of requests sent to the LLM validator by approximately 50\%  while preserving accuracy.
    \item Our proposed method establishes the state-of-the-art in detecting anomalies on the ComplexVAD dataset, achieving 72.5\% AUC, outperforming existing methods by 12.5\%.
\end{itemize}

The remainder of this paper is organized as follows. Section~\ref{sec:related} reviews related work in self-supervised and language-guided anomaly detection. Section~\ref{sec:method} details the proposed detect-validate architecture. Section~\ref{sec:experiments} presents experimental results and ablation studies. Finally, Section~\ref{sec:conclusion} concludes with future research directions.

\section{Related Work}\label{sec:related}
\label{sec:related_work}
VAD has gained significant attention due to its critical applications in surveillance, security, and behavior analysis \cite{Raja2023}, \cite{10.1145/3645101}, \cite{10271300}. Traditional supervised methods often require extensive labeled datasets, which are challenging to obtain given the rarity and unpredictability of anomalies or reconstruction-based methods, which assume anomalies are outliers to a learned model of normalcy \cite{9347460}, \cite{Baradaran2024}. To address these challenges, recent research has explored SSL \cite{math12050758}, \cite{9568697}, \cite{Li_2021_CVPR}, and the integration of LLMs for VAD \cite{10.1007/978-3-031-78125-4_25}, \cite{Zanella_2024_CVPR}, \cite{10.1145/3664647.3681190}.
 This section analyzes advancements and limitations in both domains.

\subsection{Self-Supervised Learning Approaches}
SSL approaches aim to learn meaningful representations from unlabeled data by formulating pretext tasks that capture the inherent structures in video sequences. These tasks enable models to understand normal patterns, facilitating the detection of deviations indicative of anomalies. 
A key milestone is the SSMTL framework \cite{georgescu2021anomaly}, which employs multi-task learning (MTL) (e.g., motion prediction, frame reconstruction) to capture diverse normal patterns and enhance the model's ability to detect anomalies without relying on labeled datasets. Following this, \cite{panariello2022consistency} introduced a consistency-based self-supervised loss to ensure anomaly score alignment across different augmentations of the same video snippet, which significantly improves temporal localization performance. \cite{BARBALAU2023103656} enhanced the original SSMTL model, naming it SSMTL++. They integrate multi-head self-attention, segmentation prediction, inpainting, pose estimation, and adversarial pseudo-anomaly learning to enrich spatiotemporal representations.
\cite{10775008} introduced StridedConv3D-VideoSwin into the SSMTL framework, combining the local spatial inductive biases of CNNs with the global modeling capabilities of transformers. By embedding 3D convolutions into the VideoSwin \cite{liu2022video} architecture, the method enhances feature extraction for video anomaly detection and achieves improved performance on multiple benchmarks while maintaining lower memory usage despite increased computational cost. \cite{WANG2026112016} proposed SC-MAFC, a semantic consistency-based framework that integrates appearance, optical flow, and pose through a three-branch encoder-decoder architecture, enforcing semantic consistency and enhancing multi-attribute feature complementarity.
 
\subsection{Language-Guided Anomaly Detection}
The advent of LLMs, MLLLMs and VLLMs has opened new avenues for VAD by enabling models to understand and reason about video content through natural language \cite{tang2025videounderstandinglargelanguage}, \cite{Xiao2025}, \cite{Zhao_2025_CVPR}. These models can generate textual descriptions of video frames eliminating the need for task-specific fine-tuning and assess their normality based on learned knowledge \cite{chen2024sharegpt4video}, \cite{tariq4812289vidcap}. For example, VideoLLM \cite{chen2023videollm} linearizes video frames into token embeddings and feeds them into frozen LLMs to perform multimodal reasoning across eight different video tasks. Video-LLaMA \cite{zhang2023video} and its successor Video-LLaMA2 \cite{cheng2024videollama2advancingspatialtemporal} fuse visual and audio encoders with LLaMA-based backbones, enabling instruction-following over spatial and temporal inputs. Building upon this trajectory, VideoLLaMA3 \cite{zhang2025videollama3frontiermultimodal} adopts a vision-centric philosophy. Rather than relying on massive low-quality video-text data, it prioritizes high-quality image-text pretraining and introduces innovations like any-resolution vision tokenization and dynamic token pruning. 

The capabilities of these models have naturally inspired applications in downstream tasks such as anomaly detection.
  A pioneering work, CLIP-AD \cite{chen2024clip}, leveraged CLIP's cross-modal alignment to score frame anomalies via handcrafted prompts (e.g., "a suspicious event"). However, this approach struggles with dynamic scenes due to its static textual queries. A significant leap came with LAVAD \cite{Zanella_2024_CVPR}, which adopts a truly training‑free, caption‑then‑reason pipeline. They generate frame captions by a BLIP‑2 \cite{li2023blip}, followed by GPT‑4-based aggregation and anomaly evaluation.
VadCLIP \cite{wu2024vadclip} explores weakly-supervised vision-language alignment to further refine anomaly detection without requiring dense labels or task-specific retraining. Building on this foundation, the same team introduced AVadCLIP \cite{wu2025avadclip}, which extends the CLIP-based framework to incorporate audio-visual collaboration for robust anomaly detection.

An emerging direction in language-guided VAD is the integration of rule-based reasoning, which enables models not only to detect deviations in raw visual features but also to interpret them through explicitly defined or induced logical structures. A pioneering work in this vein is AnomalyRuler \cite{yang2024follow}, which frames anomaly detection as a two-stage reasoning process. The architecture enables zero-shot, interpretable anomaly detection and achieves strong results across multiple standard benchmarks. Building upon this paradigm, other studies further explore rule-based and symbolic reasoning. 
SlowFastVAD \cite{ding2025slowfastvad} employs a dual-speed framework, where a fast visual anomaly detector is coupled with a retrieval-augmented LLM that reasons over ambiguous segments using external knowledge. From a symbolic angle, VED-SR \cite{zeng2025objectseventsunlockingcomplex} adopts a training-free framework that applies symbolic regression to discover logical expressions governing normal event patterns. 

Collectively, these methods reflect a growing trend toward structured, interpretable, and multimodal reasoning in VAD.

\subsection{Research Gap Analysis}
Existing approaches in VAD demonstrate complementary strengths and inherent limitations. SSL methods excel at capturing localized spatiotemporal features and low-level motion patterns; however, they often fail to capture high-level semantic context and inter-agent dependencies, achieving only 60\% AUC on the ComplexVAD benchmark. In contrast, LLM-based techniques are capable of reasoning over interactions and extracting semantic meaning, yet they typically introduce significant computational overhead. To overcome these challenges, We introduce HyCoVAD, a unified framework that employs SSL to detect frames suspected of anomaly and leverages LLM reasoning to validate and confirm these suspicious frames. This integration of SSL efficiency with LLM semantic understanding leads to a substantial performance gain on ComplexVAD, significantly surpassing existing baselines.

\section{Methodology}\label{sec:method}
To address the challenges of complex video anomaly detection, we propose a novel two-stage framework that integrates a self-supervised deep multitask architecture with a training-free LLM-based module. In the first stage, the deep multitask model scans the video and selects a subset of suspect frames that are more likely to contain anomalous events, based on low-level spatiotemporal deviations. In the second stage, a vision language model (VLM) and a LLM are combined within a specialized architecture to semantically detect and confirm the truly anomalous frames of this suspect set.

Our approach is designed to efficiently identify potential anomalies based on low-level spatiotemporal deviations and then semantically verify these proposals using high-level contextual information, significantly reducing computational overhead compared to end-to-end LLM-based methods.

\begin{figure}[H]
    \hspace*{-0.5cm}
    \centering
    \includegraphics[width=1.1\linewidth]{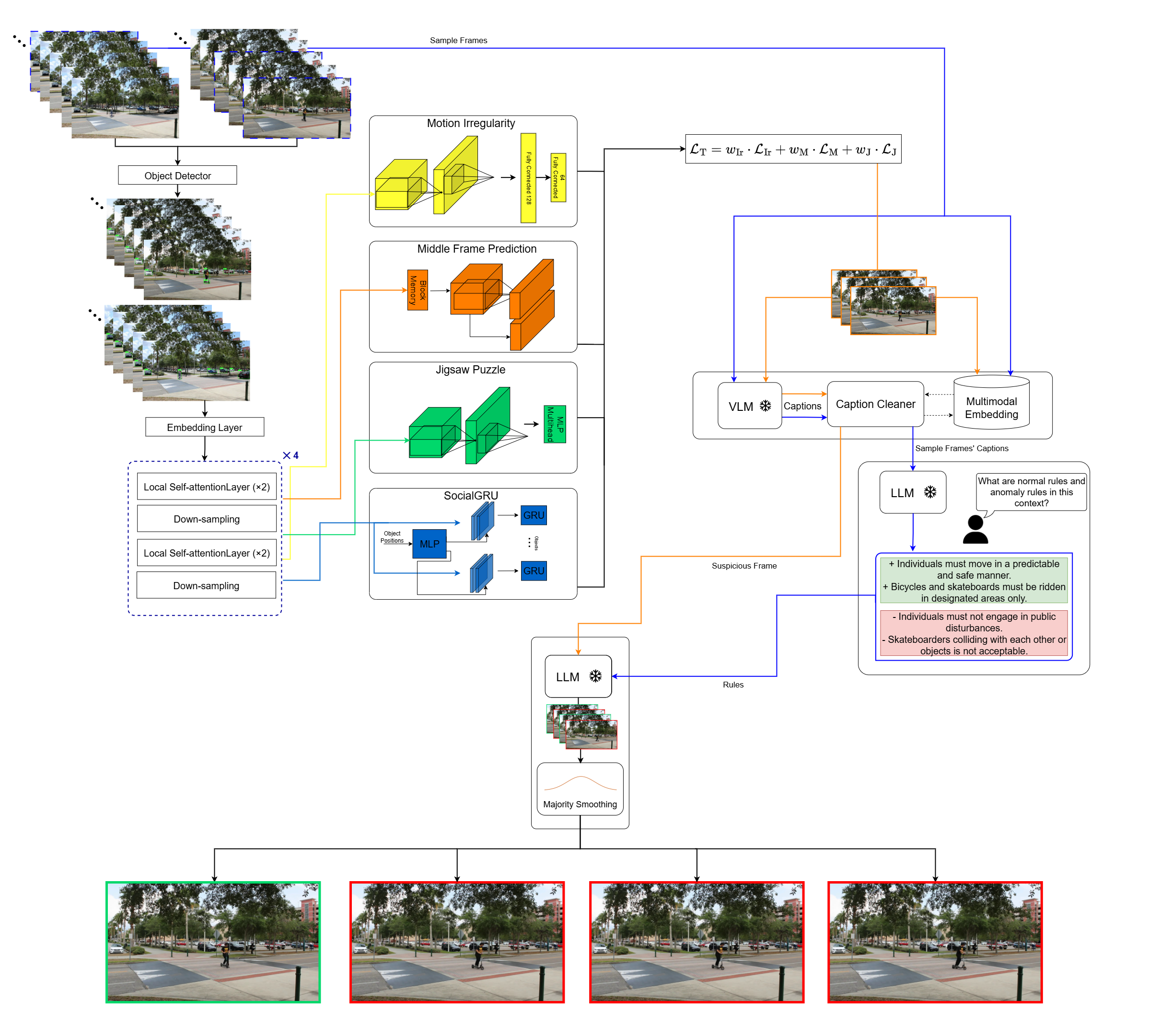}
    \caption{Overview of the proposed two-stage video anomaly detection framework. Stage 1 processes video frames at both object and frame levels. Objects are detected, and object-centric temporal sequences are extracted. A multi-task self-supervised deep model with a shared nnFormer encoder performs middle-frame prediction with memory-augmented autoencoding, motion irregularity classification, jigsaw-based spatial prediction, and SocialGRU trajectory prediction with interleaved attention. Task-specific weights are learned using GradNorm. Stage 2 performs semantic verification: frame captions are generated and refined using temporal and embedding-based alignment, and a LLM evaluates each frame against automatically generated environment-specific rules. Final predictions are smoothed with majority smoothing.\newline
    In the output examples, the first frame is detected as normal, where a scooter moves in a straight direction along the street. However, in the subsequent three frames, the scooter suddenly turns, an illegal and dangerous maneuver. This behavior is flagged as anomalous by the system, highlighting its ability to capture context-dependent rule violations.}
    \label{fig:myfigure}
\end{figure}

\subsection{Object Detection}
Our analyses are conducted at both the frame level and the object level. Considering objects individually allows the model to attend to specific entities within a scene, while frame-level processing provides a more holistic view of all objects together. As an initial step, we apply object detection on every frame using YOLOv12 \cite{tian2025yolov12attentioncentricrealtimeobject}. For each object detected in frame $i$, we construct an object-centered temporal sequence by cropping the corresponding bounding box from the neighboring frames ${i-t, \ldots, i-1, i, i+1, \ldots, i+t}$. Each cropped region is resized to $64 \times 64$ pixels. The resulting sequence forms the object-centric input for our framework.

\subsection{Multi-Task Self-Supervised Deep Learning Module}
Our deep model is built upon a MTL paradigm, where a shared visual encoder is trained jointly on multiple self-supervised tasks. This strategy allows the model to learn complementary representations from different supervisory signals. Each task is designed to capture a specific aspect of video data: temporal structure, spatial composition, or relational interaction.

\subsubsection{Shared Backbone}
Our framework is built upon a shared spatio-temporal encoder that provides common representations for all task-specific branches. We explore two encoder architectures with distinct design principles, and their performance is empirically validated in our experimental section.

As a baseline, we adopt the CNN+CvT encoder introduced by \cite{BARBALAU2023103656}, which integrates convolutional layers for local feature extraction with convolutional vision transformers (CvT) to capture global dependencies.

To further improve performance, we employ the nnFormer encoder \cite{10183842}, originally proposed for segmentation tasks in complex medical domains. We hypothesize that its hierarchical transformer design is uniquely suited for our multi-task VAD framework. Unlike single-level architectures, nnFormer progressively extracts features at multiple levels of abstraction, allowing it to jointly capture both fine-grained irregularities (e.g., subtle motion deviations) and long-range contextual patterns (e.g., complex interactions). This flexible feature sharing across tasks that rely on different scales of representation is particularly beneficial for our diverse self-supervised tasks. Furthermore, nnFormer's demonstrated robustness to noisy and heterogeneous data directly translates to the VAD problem, where subtle anomalies are often inconsistent and difficult to model. We provide a detailed ablation study in \ref{ablationStudy}, which demonstrates that the nnFormer backbone provides a performance gain over the CNN+CvT baseline, justifying its increased complexity for the task of complex anomaly detection.

\subsubsection{Middle Frame Prediction with Memory-Augmented Autoencoding}
To enhance the architecture of the middle-frame prediction task in~\cite{BARBALAU2023103656}, we incorporate the memory-augmented design of MemAE~\cite{gong2019memorizing}, which retrieves prototypical patterns from a learned memory via attention-based querying. To further improve reconstruction quality, we utilize temporal skip connections from early decoder layers, enabling the decoder to retain fine-grained spatiotemporal details.

The decoder receives both pooled encoder features and temporal skip connections, augments them through memory retrieval, expands them spatially, and reconstructs the middle frame using upsampling and convolutional layers. Skip connections at multiple resolutions are fused into the decoder pathway to improve structural accuracy.

The reconstruction loss is defined as:
\begin{equation}
\mathcal{L}_{\text{middle}} = \| I_t - \hat{I}_t \|_1
\end{equation}
where \( I_t \) is the ground-truth middle frame and \( \hat{I}_t \) is the predicted frame. This task promotes temporal consistency and enhances the model's sensitivity to subtle anomalies.

\subsubsection{Motion Irregularity}

To capture temporal consistency, we train a lightweight classifier to distinguish between regular and irregular motion in short frame sequences~\cite{BARBALAU2023103656}. The model receives an intermediate feature map and processes it through two convolutional blocks with residual connections. The flattened features are passed through two fully connected layers to produce binary predictions.
This task encourages the encoder to be sensitive to motion dynamics and discontinuities, such as harsh movements.

We use the cross-entropy loss:
\begin{equation}
\mathcal{L}_{\mathrm{irreg}} = -\frac{1}{N} \sum_{i=1}^{N} y^{(i)} \log \hat{y}^{(i)},
\end{equation}
where \(y^{(i)}\) is the regular / irregular ground-truth label and \(\hat{y}^{(i)}\) is the predicted class probability.

\subsubsection{Jigsaw Puzzle (Spatial Prediction)}

To improve spatial understanding, the model is trained to solve a \textit{jigsaw puzzle} task: an object is divided into 2 × 2 shuffled patches and the model predicts the original position of each. This is formulated as \(4\) independent multi-class classification problems, one per patch, where each patch must be assigned to its correct spatial location. This task encourages learning of object structures and spatial layout.

The decoder head processes encoder features through a convolutional block followed by adaptive average pooling, reducing the feature map to a \( 2 \times 2 \) grid. Each resulting patch feature vector is then passed through a shared two-layer MLP head consisting of a linear projection to an intermediate feature space, a ReLU activation, dropout regularization, and a final linear layer that outputs a classification over the original patch positions.

The training objective is:
\begin{equation}
\mathcal{L}_{\text{s-jigsaw}} = -\frac{1}{4} \sum_{i=1}^{4} \log \hat{y}_{i, y_i},
\end{equation}
where \(y_i\) is the true position index of patch \(i\), and \(\hat{y}_{i, y_i}\) is the predicted probability for the correct class. 

\subsubsection{SocialGRU}

Considering the path of moving objects, this task enables us to monitor their positions and verify whether they follow a normal trajectory. Since the behavior of a moving object changes with respect to nearby objects in proximity, we propose a network that tracks the path of each object relative to the paths of neighboring objects.

We employ a lightweight GRU unit both as encoder and decoder \cite{Alahi_2016_CVPR}, combined with an interleaved attention mechanism. This attention network is trained jointly with the GRUs to fuse the behavior of neighboring objects into the representation of the target object. The attention weights allow the model to focus on the most influential neighbors by assigning them higher scores.

To improve the Social GRU, we additionally concatenate the hidden state of the encoder with the object feature vector extracted over the past 7 frames. This augmented representation provides the model with explicit object identity information, enabling it to relate past trajectories to object-specific features.
Our model is trained to predict the position of each object at the 8th frame given its positions over the previous 7 frames, by minimizing the mean squared error (MSE) loss:
\begin{equation}
  \mathrm{MSE} \;=\; \sum_{i=1}^{N} \bigl\| y_i - \hat{y}_i \bigr\|^2
\end{equation}
where \(y_i\) is the ground-truth position and \(\hat{y}_i\) is the predicted position of object \(i\).

\subsubsection{Task-Specific Weight Learning}

Multitask learning often suffers from imbalanced task optimization, where some tasks dominate training, while others lag. To address this, we adopt \textit{GradNorm} \cite{chen2018gradnorm}, which dynamically adjusts task weights by aligning their gradient magnitudes to balance training rates.

The total loss is defined as:
\begin{equation}
\mathcal{L}_{\text{total}} = \sum_{i=1}^{T} w_i(t) \cdot \mathcal{L}_i(t)
\end{equation}
where \( w_i(t) \) are task-specific weights learned during training. GradNorm updates these weights based on each task's relative training speed, computed using the normalized loss ratio \( \tilde{L}_i(t) = \mathcal{L}_i(t)/\mathcal{L}_i(0) \). The target gradient norm for each task is:
\begin{equation}
G^{\text{target}}_i(t) = \bar{G}(t) \cdot \left( \frac{\tilde{L}_i(t)}{\frac{1}{T} \sum_{j=1}^{T} \tilde{L}_j(t)} \right)^{\alpha}
\end{equation}
where \( \bar{G}(t) \) is the average gradient norm and \( \alpha \) controls the degree of balance.

The learned weights \( w_i \) are used for inference, which ensures consistency between training and anomaly scoring.

\subsection{LLM-Based Semantic Reasoning and Anomaly Verification}

The process begins with frame-level captioning using a pretrained VLM, which generates descriptive textual summaries of visual content. To ensure alignment between visual information and generated descriptions, we apply a caption refinement step based on multimodal embedding similarity, enhancing semantic fidelity for challenging frames (e.g., those with blur, occlusion, or ambiguity).
Next, to provide the LLM with contextual grounding, we sample a set of unlabeled frames' captions from the test dataset and prompt the LLM to infer environment-specific norms by analyzing their captions. This allows the model to generate a set of context-aware rules that distinguish normal from abnormal behavior.
Finally, each suspect frame's refined caption and the extracted contextual rules are presented to the LLM. Through a structured prompt, the LLM performs a verification task to determine whether the described scene constitutes an anomaly.

By explicitly integrating the LLM's decision-making in rule sets along with semantic frame captions, this approach reduces ambiguity and improves interpretability compared to caption-only input. The structured prompt ensures consistent, rule-based reasoning that can generalize across scenes.

\subsubsection{Caption Generation}

To enable semantic reasoning via an LLM over visual content, we convert video frames into natural language descriptions through a captioning stage. This process abstracts low-level pixel data into textual representations that describe human activities, object interactions, and scene contexts relevant for anomaly detection. 

\subsubsection{Caption Refinement} \label{sebsec:clean}

Automated frame-level captions in video-language datasets often exhibit noise or weak semantic alignment with visual content, largely due to motion blur, occlusions, or captioning errors. To address this, we employ a caption refinement strategy that enhances temporal and visual consistency using a joint embedding space.

To improve semantic alignment, each original suspect frame's caption is refined by selecting the most similar alternative from a local temporal window based on cosine similarity to the frame's visual embedding, demonstrated as

\begin{equation}
\hat{c}_t = \arg\max_{c_j \in \mathcal{W}_t} \, \text{Sim}( \text{Embed}_{\text{img}}(f_t), \text{Embed}_{\text{txt}}(c_j)),
\end{equation}
 where $\hat{c}_t$ is the refined caption selected for frame~$t$, 
$\mathcal{W}_t$ denotes the local temporal window of candidate captions around frame~$t$, 
$f_t$ represents the visual content of frame~$t$, 
and $c_j$ is a caption candidate within $\mathcal{W}_t$.
To maintain lightweight preprocessing while fully utilizing all caption content, we adopt a sentence-level chunking strategy optimized for small text encoders. Given a caption composed of \( N \) sentences \( \{ s_1, s_2, \dots, s_N \} \), we construct overlapping chunks by concatenating \( m \) consecutive sentences with an overlap of \( o \) sentences between chunks, where \( 1 \leq o < m \). This approach produces semantically rich but compact textual segments, enabling robust encoding under the constraints of lightweight embedding models.

The $k$-th chunk is defined as:
\begin{equation}
\begin{aligned}
c_k &= s_{k} \, \Vert \, s_{k+1} \, \Vert \, \dots \, \Vert \, s_{k + m - 1} \\
    &\text{for} \quad k = 1,\, 1 + (m - o),\, 1 + 2(m - o),\, \dots
\end{aligned}
\end{equation}

The total number of chunks (demonstrated by $K$) is given by:

\begin{equation}
K = \left\lfloor \frac{N - m}{m - o} \right\rfloor + 1
\end{equation}

Each chunk \( c_k \), taken from the same caption, is independently processed using the text encoder \( \text{Embed}_{\text{txt}}(\cdot) \). This maps each chunk into a shared latent space:
\begin{equation}
z_k = \text{Embed}_{\text{txt}}(c_k), \quad z_k \in \mathbb{R}^d.
\end{equation}

where \( d \) is the dimension of the shared latent space. To obtain a single representation for the entire caption, the embeddings of all its chunks are aggregated. Here, $\text{Pool}(\cdot)$ is an aggregation function used to combine chunk embeddings, commonly implemented as mean or max pooling. The resulting caption-level embedding is given by Eq.~\ref{eq:pool}.

\begin{equation}
z_{\text{caption}} = \text{Pool}( z_1, z_2, \dots, z_K )
\label{eq:pool}
\end{equation}

We have explored both \textit{local} (temporal window) and \textit{global} (full video) modes. The local search prioritizes temporal coherence, while the global search maximizes semantic alignment.
This refinement process improves temporal coherence and semantic fidelity in downstream tasks.

\subsubsection{Rule Generation}

Our framework incorporates an unsupervised rule generation module that uses LLM to derive general, context-aware rules for distinguishing normal from anomalous behaviors in video data, based on precomputed frame-level captions.

Given the high frame count per video, analyzing every caption is inefficient and often unnecessary. To address this, we sample 20 frames per video so that representative behaviors are captured efficiently, while ensuring temporal and contextual diversity.
Each sampled caption is processed with structured prompts designed to guide the LLM in generating precise, environment-specific rules. These rules describe normative behaviors (e.g., orderly pedestrian flow, traffic compliance) and potential anomalies (e.g., sudden movements, unexpected actions, obstructions).

Since rule generation occurs per frame, outputs may contain redundancy or contradictions. We address this via an aggregation step, consolidating similar rules, resolving conflicts, and enforcing consistent formatting to create a unified rule set per video. Normal and anomaly rules are separated clearly, duplicates are removed, and all unique anomaly cases are retained. In the presence of contradictory rules, the aggregation follows a worst-case principle, adopting the stricter rule to ensure anomalies are not overlooked. This guarantees both completeness and robustness in the final rule set. Collectively, these rule sets provide explicit semantic grounding for the anomaly verification stage.

\subsubsection{Anomaly Detection via LLM}

The final anomaly detection stage uses the LLM to decide over the refined frame captions and corresponding rule sets that define normal and anomalous behaviors. For each suspect frame \( f_t \), with caption \( c_t \), we provide the LLM with explicit rules for normality and anomaly, denoted as:
\begin{equation}
\mathcal{R}_{\text{normal}} = \{ r^{(n)}_1, r^{(n)}_2, \dots \}, \quad
\mathcal{R}\
_{\text{anomaly}} = \{ r^{(a)}_1, r^{(a)}_2, \dots \}.
\end{equation}

The LLM is queried using a structured prompt designed to enforce precise and interpretable anomaly detection. The prompt instructs the LLM to analyze the scene using only the provided rules and description, and to output whether the scene is anomalous, along with which rules are broken or preserved. The LLM output for frame \( f_t \) is represented as Eq.~\ref{eq:llm_out},
\begin{equation}
y_t = \text{LLM}\big(c_t, \mathcal{R}_{\text{normal}}, \mathcal{R}_{\text{anomaly}}\big),
\label{eq:llm_out}
\end{equation}
where \( y_t \) contains a detection of anomaly status and an explanation referencing the rules.

\subsection{Exponential Majority Smoothing} \label{sec:EMS}

To obtain stable binary predictions and compute AUC scores, we adopt \textit{Exponential Majority Smoothing (EMS)} as a post-processing step, following~\cite{yang2024anomalyruler}. EMS combines Exponential Moving Average (EMA) with local Majority Voting to reduce temporal noise in raw frame-level binary predictions. This post-processing is used during evaluation to enhance prediction consistency.

\section{Experiments}\label{sec:experiments}

This section compares our model with baselines and state-of-the-art methods in terms of detecting anomalies. We conduct an extensive ablation study to justify our main design choices regarding the proposed components, SSL module hyperparameters, prompt design, and LLM and VLM setups. Examples of complete prompts, derived rules, and outputs are illustrated in Appendix \ref{apendix:prompt}.

\subsection{Experimental Setup}

\textbf{Dataset.}
We evaluated our model using the ComplexVAD dataset \cite{mumcu2025complexvad}, a recently introduced large-scale benchmark designed to address the limitations of existing VAD datasets. Unlike prior datasets that primarily focus on simple anomalies involving single-object deviations, ComplexVAD emphasizes \textit{interaction-based anomalies}, where the anomalous behavior arises from the relationships between multiple objects or agents.

The ComplexVAD dataset comprises a total of 217 video sequences, captured at resolution of 1920×1080 pixels over various times of the day and days of the week, using a static camera overlooking a two-lane street with adjacent sidewalks and a crosswalk on the campus of the University of South Florida. The dataset is divided into:

\begin{itemize}
    \item \textbf{Training Set:} 104 videos containing only normal activities.
    \item \textbf{Testing Set:} 113 videos, which include both normal and anomalous events.
\end{itemize}

A distinctive feature of ComplexVAD is its focus on \textit{complex anomalies} that result from interactions between objects. Examples include:

\begin{itemize}
    \item A person leaving a package unattended on the sidewalk.
    \item Two individuals engaging in a physical altercation.
    \item A skateboard moving without a rider.
\end{itemize}
These scenarios are challenging because the individual objects (e.g., person, package, skateboard) are common and not inherently anomalous. It is the context and interaction between these objects that constitute the anomaly. This necessitates models to understand not just object presence but also their relationships and interactions over time.

 For the full procedure, given that the dataset contains nearly 1,800,000 frames,making exhaustive processing computationally infeasible, we sampled every 10\textsuperscript{th} frame from the test videos, resulting in approximately \textbf{180,000} being processed.

\textbf{Evaluation Metrics.} Following standard practice, we primarily report the Area Under the Receiver Operating Characteristic Curve (AUC) to evaluate detection performance. For LLM-based methods, which inherently produce binary (normal/abnormal) predictions rather than continuous anomaly scores, we convert their outputs into calibrated confidence scores using the approach described in Section~\ref{sec:EMS}. This allows us to compute AUC fairly across all models. Additionally, we report precision, recall, F1-Score, and ROC-AUC to compare binary classification performance by the SSL module. All metrics are computed at the frame level using the provided ground truth annotations.

\textbf{Implementation Details.} All experiments were conducted as part of a unified pipeline, using 10 NVIDIA RTX 4090 GPUs (24GB VRAM each) and a total of 150GB of system RAM. The training and inference pipelines were developed using PyTorch 2.0 with CUDA 12.1. We utilized \textit{YOLOv12n} model for object detection. For the LLM component of our system, we deployed \textit{{LLaMA 3.3}} with 70 billion parameters. Please refer to Appendix~\ref{apendix:llm} for a detailed analysis of the variant. To generate captions for the frame, we used \textit{LLaVA-NeXT-Video-7B-hf} \cite{liu2024llavanext} \cite{zhang2024llavanextvideo} model in 32 bit Float mode and \textit{ImageBind} \cite{girdhar2023imagebind} embedding model was leveraged in our Caption Refinement module. 

\subsection{Training Procedure}
The training phase is dedicated solely to the SSL component of our pipeline. We adopt a structured multi-phase training scheme. Since our model involves multiple proxy tasks, we employ task-specific loss functions: the {L2 loss} is used for the reconstruction-based Middle Frame Prediction task, and the {cross-entropy loss} is applied to classification tasks such as Irregularity Detection and both Jigsaw tasks. Optimization is performed using the {AdamW} optimizer, with learning rate decaying by a factor of 0.1 every 10 epochs.

The training proceeds in three progressive phases:
\begin{itemize}
    \item \textbf{Phase 0 (Encoder Only)}: The shared encoder is trained while keeping all decoders frozen, as proposed in \cite{brock2017freezeout}.
    \item \textbf{Phase 1 (Decoder Unfreezing)}: Decoder layers are gradually unfrozen, beginning from the output and moving inward layer by layer over successive epochs \cite{howard2018universal}.
    \item \textbf{Phase 2 (Full Model Training)}: The entire network, including the encoder and all decoders, is trained jointly. The optimizer is reinitialized and a unified learning rate is applied across the model.
\end{itemize}
This progressive strategy functions as a form of architectural curriculum learning. It helps establish strong shared representations before introducing the full task-specific learning burden, which is crucial for avoiding negative transfer and ensuring stable convergence in multi-task settings.

\subsection{Inference Efficiency}
After completing training, we proceed to the inference phase, where the full pipeline operates in a two-stage manner.

First, each input frame is passed through the trained SSL module, which acts as a coarse anomaly detector. Frames are flagged as potentially abnormal if their total computed loss from the SSL tasks exceeds a fixed threshold of 0.3. These flagged frames are then forwarded to the LLM-based component.
This component performs captioning, semantic alignment, rule induction, reasoning, and smoothing to determine final anomaly scores for the selected frames.

The training-free stage leverages large pretrained models for rule-based inference, yielding interpretable outputs grounded in semantic understanding. Importantly, by introducing this coarse-to-fine filtering design, we reduced the number of requests sent to the LLM by approximately 50\%, while still maintaining strong anomaly detection performance. This highlights the efficiency gains of the proposed pipeline without sacrificing accuracy.

\subsection{Comparison with State-of-the-Art}

The authors of ComplexVAD \cite{mumcu2025complexvad}, alongside releasing the dataset, introduced three baseline methods evaluated on frame-level anomaly detection. Table~\ref{tab:sota_complexVAD} presents a comparison of our method against these baselines. To the best of our knowledge, these remain the only publicly reported results on ComplexVAD due to the dataset's recent publication.

\begin{table}[h]
\centering
\caption{Comparison of frame-level AUC scores on the ComplexVAD dataset.}
\label{tab:sota_complexVAD}
\begin{tabular}{lc}
\toprule
\textbf{Method} & \textbf{Frame-Level AUC (\%)} \\
\midrule
\rowcolor{red!10}MemAE \cite{gong2019memorizing} & 58 \\
\rowcolor{red!10}EVAL \cite{singh2023eval} & 54 \\
\rowcolor{red!10}ComplexVAD (Graph-Based) \cite{mumcu2025complexvad} & 60 \\
\rowcolor{green!10} {HyCoVAD (Ours)} & {72.5} \\
\bottomrule
\end{tabular}
\end{table}

HyCoVAD achieves a substantial performance gain over all prior methods on the ComplexVAD benchmark, with a frame-level AUC of \textbf{72.5\%}, compared to 60.0\% by the best baseline. This represents a \textbf{12.5\% absolute improvement} over the strongest reported result and demonstrates the potential of incorporating LLMs into the VAD pipeline, especially for complex semantic understanding.

Unlike traditional methods such as MemAE or EVAL that rely solely on low-level reconstruction errors or weak supervision, HyCoVAD benefits from a hybrid pipeline: SSL module handle frame-level perception, while LLM-based component refine decisions based on high-level contextual consistency. This combination helps reduce false positives and improves robustness across challenging scenes.

While our approach includes a rule-based judgment module driven by LLMs, our primary goal is not to contribute to the domain of reasoning per se. Since ComplexVAD lacks explicit annotations for reasoning or explanation tasks, we do not formally evaluate reasoning accuracy. Nevertheless, the rule-based mechanism enables our model to identify semantically anomalous events that are often missed by conventional visual-only systems. Taken together, these results indicate that our method enables both strong detection performance and interpretable insights in complex video environments. 

\subsection{Ablation Study}\label{ablationStudy}

We first evaluated different configurations for the SSL-based anomaly detector. We explored various combinations of architectures, input modalities, and preprocessing techniques to determine the most effective setup for detecting anomalies directly from video data, summarized in Table~\ref{tab:ablation_deep_models}.
\begin{table*}[t]
\centering
\caption{Performance of the SSL module across different setups}
\label{tab:ablation_deep_models}
\resizebox{\textwidth}{!}{
\begin{tabular}{l c c c c c c}
\hline
\textbf{Backbone} & \textbf{Tasks} & \textbf{Level} & \textbf{Recall (\%)} & \textbf{Precision (\%)} & \textbf{F1-Score (\%)} & \textbf{ROC-AUC (\%)} \\
\hline
\multirow{4}{*}{3D CNN + CVT} 
  & \multirow{2}{*}{T1 + T2 + T3 + T4} & Object & 53 & 50 & 51.45 & 52.41 \\
  &                                    & Frame  & 49 & 46 & 47.5 & 51.17 \\
\cmidrule{2-7}
  & \multirow{2}{*}{T1 + T2 + T3}      & Object & 50 & 47 & 48.5 & 51.08 \\
  &                                    & Frame  & 47 & 44 & 45.5 & 50.8\\
\midrule
\multirow{4}{*}{NNFormer}
  & \multirow{2}{*}{T1 + T2 + T3 + T4} & Object & 56 & 54 & {55} & \cellcolor{green!15} \textbf{58.32}\\
  &                                    & Frame  & 53 & 51 & 52 & 54.07 \\
\cmidrule{2-7}
  & \multirow{2}{*}{T1 + T2 + T3}      & Object & 53 & 51 & 52 & 52.49 \\
  &                                    & Frame  & 50 & 47 & 48.5 & 51.35 \\
\bottomrule
\end{tabular}
}
{\centering\footnotesize\itshape \newline \newline T1: Motion Irregularity, T2: Middle Frame Prediction, T3: Spatial Jigsaw, T4: Social GRU\par}
\end{table*}

After identifying the best-performing SSL module configuration, we proceeded to evaluate and refine the second stage of our architecture. To better understand the contributions of each component and design choice in our LLM-based anomaly detection pipeline, we conducted a comprehensive ablation study. We divide our analysis into two main categories: (1) component-wise alterations or removals, and (2) variations in key hyperparameters. Each set of ablations isolates specific factors to evaluate their individual impact on performance, as measured by frame-level AUC.
    
Importantly, adding the LLM stage increases the AUC by approximately 12.18\%, demonstrating the clear need and effectiveness of this stage in complementing the SSL module.
\subsubsection{Component-Level Modifications}
In this set of experiments, we examined the effect of removing or replacing individual components of our pipeline. Removing the rule judgment stage entirely, testing a CLIP2-based captioning model, skipping the cleaning process, and adding a memory mechanism that carried forward anomaly flags to subsequent frames were our component-level modifications, which their results are presented in Table~\ref{tab:ablation_components}. Interestingly, adding short-term memory substantially harmed performance. Once an anomaly was detected, subsequent frames tended to be mislabeled as anomalous due to memory propagation, even though anomalies in most videos occur only briefly. This false continuation effect reduced overall performance.
Furthermore, these modifications show the critical role of the rule component, as when the rule stage is removed, performance drops sharply to 59.2\%. This reduction demonstrates that rule-based filtering is indispensable for ensuring reliable anomaly detection.

\begin{table}[h]
\centering
\caption{Component Ablations and Performance}
\begin{tabular}{lcc}
\toprule
\textbf{Ablation Setting} & \textbf{Component Change} & \textbf{AUC (\%)} \\
\midrule
No Rule-Based Filtering & Rule component removed & 59.2 \\
CLIP2 Captioner & Replaced captioner model & 60.7 \\
No Cleaning & Cleaning component removed & 64.6 \\
With Short-Term Memory & Added memory & 56.1 \\
\bottomrule
\end{tabular}
\label{tab:ablation_components}
\end{table}

\subsubsection{Hyperparameter Variations}
This section explores the effects of tuning different hyperparameters within our architecture, which are provided in Table~\ref{tab:ablation_hyperparameters}. Increasing the number of caption tokens generally improved performance up to a certain point. The best result was observed at 200 tokens, while further increasing to 300 tokens led to a slight decrease in performance. We compared window sizes of 10, 15, and 20. A size of 10 yielded the best result, compared to 65.8\% (window=15) and 66.3\% (window=20).
In addition, captions generated with Float32 precision slightly outperformed Float16.

\begin{table}[h]
\centering
\caption{Hyperparameter Ablation Results}
\begin{tabular}{lcc}
\toprule
\textbf{Hyperparameter Setting} & \textbf{Value / Variant} & \textbf{AUC (\%)} \\
\midrule
\multirow{4}{*}{Caption Token Length} 
 & 100 & 63.9 \\
 & 150 & 67.4 \\
 & 200 & \textbf{72.5} \\
 & 300 & 68.1 \\
\midrule
\multirow{3}{*}{Cleaning Window Size} 
 & 10 & \textbf{72.5} \\
& 15 & 65.8 \\
 & 20 & 66.3 \\
\midrule
\multirow{2}{*}{Captioner Precision} 
 & float16 & 62.8 \\
 & float32 & \textbf{72.5} \\
\bottomrule
\end{tabular}
\label{tab:ablation_hyperparameters}
\end{table}

Overall, these results confirm that the integration of rule-based reasoning, robust caption cleaning, memory mechanisms, and carefully tuned hyperparameters significantly improve the performance of anomaly detection in our LLM-based pipeline.

\section{Conclusion} \label{sec:conclusion}

In this work, we introduced HyCoVAD, a novel hybrid framework designed to address the challenges of complex, interaction-driven video anomaly detection. Our approach integrates the efficiency of a SSL module for generating anomaly proposals with the semantic reasoning capabilities of a LLM for validation. Specifically, HyCoVAD employs an nnFormer-based multi-task SSL temporal analyzer to efficiently identify frames exhibiting low-level spatiotemporal irregularities. These suspicious frames are then subjected to a semantic verification stage, where the LLM leverages auto-generated, context-specific rules to confirm high-level, interaction-based anomalies.
Extensive experiments on the challenging ComplexVAD dataset demonstrate the effectiveness of this detect-validate strategy. HyCoVAD achieves a state-of-the-art frame-level AUC of 72.5\%, outperforming existing baselines, while also reducing the computational overhead associated with frame-by-frame LLM analysis.

Future work can explore several promising research directions. First, extending HyCoVAD to a multimodal setting by incorporating audio signals could further enrich anomaly detection, as many real world anomalies, such as verbal disputes, carry strong auditory cues. Second, while our coarse-to-fine filtering mechanism already reduces LLM usage, additional efficiency gains may be achieved through knowledge distillation, transferring reasoning capabilities from the large LLM into a smaller, task-specific model for anomaly verification. Moreover, developing a more dynamic and adaptive rule generation process that evolves throughout a video sequence could further strengthen semantic reasoning. Finally, evaluating HyCoVAD's generalizability across broader datasets with diverse and complex anomalies would provide a stronger foundation for validating the potential of hybrid SSL-LLM architectures in the video anomaly detection domain.
\newline
\newline
\noindent{\textbf{Data Availability.}} All the data used for this research is publicly accessible and proper citations are also provided in the article. The source code that supports the findings of this study is available from the corresponding author upon reasonable request.

\bibliography{sn-bibliography}

\begin{appendices}

\section{Large Language Models Used} \label{apendix:llm}
In this section, we list all LLMs evaluated during the development of our anomaly detection pipeline, presented in Table~\ref{tab:llms}. 
Each model is characterized by its parameter count, quantization scheme, and performance score. 
All models, except for \textit{Qwen3-32B} which was served using \textit{vLLM}, were deployed via \textit{OLLAMA} on our dedicated server.

\begin{table}[h!]
\centering
\caption{LLMs tested in the pipeline. The final chosen model is highlighted.}
\renewcommand{\arraystretch}{1.2}
\begin{tabular}{lccc}
\hline
\textbf{Name} & \textbf{Parameters} & \textbf{Quantization} & \textbf{Performance (\%AUC)} \\
\hline
DeepSeek-R1 & 14B & - & 63.4 \\
Qwen3 & 32B & - & 67.5 \\
Phi-4 & 14B & FP16 & 61.8 \\
\rowcolor{green!15}
LLaMA3.3 & 70B & Instruct-Q2\_K & \textbf{72.5} \\
\hline
\end{tabular}

\label{tab:llms}
\end{table}

\section{Vision Language Models Used}\label{apendix:vlm}

In order to determine the most suitable VLM for producing frame descriptions, we carried out a targeted qualitative assessment of three candidates: \textbf{BLIP-2}, \textbf{LLaVA-NeXT-Video-7B-hf}, and \textbf{LLaVA-NeXT-Video-34B-hf}. 
The selection criteria were driven by three main factors: accuracy in describing salient scene elements, brevity without loss of essential information, and practicality in terms of computational overhead.  
This evaluation was performed using a representative sample of frames from the \textit{ComplexVAD} dataset, covering both normal and anomalous events. Results are summarzied in Table~\ref{tab:vlm_comparison}
\begin{table}[h!]
\centering
\caption{Summary of candidate VLMs for caption generation on the \textit{ComplexVAD} dataset. The final selected model is highlighted.}
\renewcommand{\arraystretch}{1.3}
\begin{tabular}{p{3.9cm} p{9cm}}
\hline
\textbf{Model} & \textbf{Summary \& Decision} \\
\hline
BLIP-2 & Fast and reliable in identifying primary objects, but produced captions with insufficient detail for complex anomaly detection. Lacked coverage of key interactions, leading to its exclusion. \\
\hline
LLaVA-NeXT-Video-34B-hf & Generated slightly more detailed captions than the 7B version but suffered from very long inference times and excessive verbosity, making large-scale processing of frames impractical. \\
\hline
\rowcolor{green!10}
LLaVA-NeXT-Video-7B-hf & Achieved a practical balance of detail and runtime efficiency. While slightly less descriptive than 34B, it maintained good performance and was computationally feasible for processing the entire dataset, leading to its selection. \\
\hline
\end{tabular}
\label{tab:vlm_comparison}
\end{table}

\section{LLM Prompting Walkthrough}\label{apendix:prompt}
In this section, we document the exact prompts and example outputs used in each stage of our LLM-based anomaly detection pipeline. This ensures reproducibility and provides insight into how the model interprets frames, generates rules, and makes final decisions.
Note that Caption Refinement module, as described in Sec.~\ref{sebsec:clean}, doesn't use any prompts, therefore is not mentioned here.

\subsection{Frame Caption Generation}
At time step $t$, the VLM receives an instruction $I_G$ to describe the frame $f_t$, generating caption $C_t$. These captions are refined, denoted by $\hat{C}_t$, and are used in final decision making.

\noindent\textbf{Instruction to VLM Model:}

\promptbox[yellow!10]{
$I_G$ = You are an AI visual assistant that can analyze a video frame. Describe the frame with focus on people's activities, objects and their relation. Don't mention background.
}

\noindent\textbf{Example Output:}

\promptbox[green!10]{
$C_t$ = In the image, there is a person in the center of the frame who appears to be in motion, possibly walking or riding a scooter, as they are captured mid-stride. The individual is wearing a light-colored shirt and dark pants, and they are positioned on a sidewalk adjacent to a crosswalk. The person is facing away from the camera, suggesting they are moving towards the right side of the frame.
\newline
On the left side of the frame, there is a car parked on the side of the road, and a person is standing near it, possibly waiting or observing the movement of the individual in the center. The car is parked on the side of the road, and the person is standing close to it, possibly indicating they are waiting for the individual to pass.
On the right side of the frame, there is a car parked on the side of the road, and a person is standing.
}

\subsection{Rule Generation}
Given the caption $C_t$, the LLM is prompted with $P_R$ to produce a set of task-specific rules for the sampled frame. After generating rule sets for all sampled frames of one video, rules are concatenated and aggregated by prompting the LLM with $P_A$, to form a concise, uniform set of rules $R_A$ for the video.

\textbf{Prompt $P_R$ for Rule Generation:}

\promptbox[yellow!10]{
$P_R$ = You are an expert in analyzing scenes and generating rules for normal and anomaly detection. Now given the following frame description:
\newline
[{frame description $C_t$}]
\newline
\newline
Generate a set of rules that define:
\newline
1. What constitutes a normal activity or object in this context.
\newline
2. What constitutes an anomaly activity or object in this context.
\newline
\newline
The only anomalies that matter are interactions of two or more objects/actors. Be strict, there should not be anything odd or out of ordinary in the street or on the sidewalk of the campus. Adhere to these notes.
Rules should be clear, concise, and specific to the context of the frame description and the needed anomalies we said. Don't miss any interacted objects.
}

\textbf{Prompt $P_A$ for Rules Aggregation}

\promptbox[yellow!10]{
        $P_A$ = You are an expert at organizing rules for video anomaly detection. Below are multiple independent rule sets 
        for the same video. Combine them into one final set of rules with:
        \newline
            1. Clear separation between normal and anomaly rules
        \newline
            2. No duplicate rules
        \newline
            3. All unique anomaly cases preserved
        \newline
            4. Contradictory rules resolved by considering the worst case (apply stricter rules)
        \newline
            5. Consistent formatting, with no markdown annotations, no extra talks
        \newline
        \newline
        Rules to aggregate:
        \newline
            [all generated rules for this video]
        \newline
        \newline
        Provide the consolidated rules in this exact format. Don't output anything more than the given format:
        \newline
        \newline
            === RULES FOR NORMAL ACTIVITIES OR OBJECTS ===
        \newline
            - [normal rule 1]
        \newline
            - [normal rule 2]
        \newline
            ...
        \newline
        \newline
            === RULES FOR ANOMALY ACTIVITIES OR OBJECTS ===
        \newline
            - [anomaly rule 1]
        \newline
            - [anomaly rule 2]
        \newline
            ...
}

\textbf{Rule Set Output $R_A$:}

\promptbox[green!10]{
$R_A$ = \#\#\# RULES FOR NORMAL ACTIVITIES OR OBJECTS
\newline
- Vehicles and pedestrians must adhere to traffic signals and pedestrian crossings.
\newline
- Skateboarders and bikers are allowed in designated areas with proper safety gear.
\newline
- Golf carts are permitted in golf courses and must be used according to the course rules.
\newline
- Trolleys are to be used within store premises by authorized personnel.
\newline
\newline
\#\#\# RULES FOR ANOMALY ACTIVITIES OR OBJECTS
\newline
- Vehicles must not exceed speed limits or drive recklessly.
\newline
- Pedestrians should not jaywalk or obstruct traffic flow.
\newline
- Skateboarders and bikers must avoid colliding with others or stationary objects.
\newline
- Climbing poles without authorization is prohibited for safety reasons.
}

\subsection{Anomaly Detection}
The decision-making step uses a structured template prompt $P_D$ that combines video's aggregated rules $\mathcal{R}_A$ with the its current refined caption $\hat{C}_t$. Finally, the LLM produces an anomaly decision output $O_t$ given $\tilde{P}_D$ prompt.

\textbf{Prompt $P_D$ for Decision Making:}

\promptbox[yellow!10]{
$R_t$ =  You are an expert in anomaly detection in video frames. You should be very strict and don't let any anomalies or broken rules unattended. Use scene description and rules provided. Below is t scene description and the video's rules:
\newline
\newline
Scene Description:
\newline 
[frame's refined caption $\hat{C}_t$]
\newline 
\newline 
Rules for Normal/Anomaly Scenes:
\newline 
[video's aggregated rules $R_A$]            
\newline 
\newline 
Analyze this scene and determine:
\newline 
            1. Is this an anomaly scene? (Answer Yes/No first)
\newline 
            2. If anomaly, which rules are broken?
\newline 
\newline 
Format your response like this:
\newline 
            Anomaly: Yes/No
\newline 
            Broken Rules: [list broken rules or "None"]
}

\noindent\textbf{Example Completed Prompt $\hat{P}_D$:}

\promptbox[orange!10]{
 $\hat{P}_D$ = You are an expert in anomaly detection in video frames. You should be very strict and don't let any anomalies or broken rules unattended. Use scene description and rules provided. Below is a scene description and the video's rules:
            \newline
            \newline
Scene Description:
            \newline
    In the image, there is a person in the center of the frame who appears to be in motion, possibly walking or riding a scooter, as they are captured mid-stride. The individual is wearing a light-colored shirt and dark pants, and they are positioned on a sidewalk adjacent to a crosswalk. The person is facing away from the camera, suggesting they are moving towards the right side of the frame.
    On the left side of the frame, there is a car parked on the side of the road, and a person is standing near it, possibly waiting or observing the movement of the individual in the center. The car is parked on the side of the road, and the person is standing close to it, possibly indicating they are waiting for the individual to pass.
    On the right side of the frame, there is a car parked on the side of the road, and a person is standing.
            \newline
            \newline
Rules for Normal/Anomaly Scenes:
            \newline
    === RULES FOR NORMAL ACTIVITIES OR OBJECTS ===
            \newline
    - Individuals must move in a predictable and safe manner.
            \newline
    - Bicycles and skateboards must be ridden in designated areas only.
            \newline
    - Vehicles must not be left in a manner that obstructs paths.
            \newline
    - Pedestrians must cross at designated times and places, following traffic signals.
            \newline
    - Cyclists and motorists must follow all traffic laws.
            \newline
            \newline
    === RULES FOR ANOMALY ACTIVITIES OR OBJECTS ===
            \newline
    - Individuals must not engage in public disturbances.
            \newline
    - Using an object as a weapon or for theft is prohibited.
            \newline
    - Skateboarders colliding with each other or objects is not acceptable.
            \newline
    - Leaving vehicles unattended in a way that obstructs traffic or pedestrian flow is forbidden.
            \newline
    - Engaging in behavior that disturbs the peace or endangers others is not allowed.
            \newline
            \newline
Analyze this scene and determine:
            \newline
            1. Is this an anomaly scene? (Answer Yes/No first)
            \newline
            2. If anomaly, which rules are broken?
            \newline
            \newline
Format your response like this:
            \newline
            Anomaly: Yes/No
            \newline
            Broken Rules: [list broken rules or "None"]
}

\textbf{Example Model Output $O_t$:}

\promptbox[green!10]{
$O_t$ = 
\newline
Anomaly: Yes
\newline
\newline
Broken Rules: 
\newline
- Individuals must move in a predictable and safe manner. (The person is riding a scooter on a crosswalk, which may not be a designated area for scooters.)
\newline
- Vehicles must not be left in a manner that obstructs paths. (Although the car is parked on the side of the road, it's not clear if it's obstructing the path or not, but the primary concern is the scooter usage on the crosswalk.)
\newline
- Individuals must not engage in public disturbances. (Riding a scooter on a crosswalk could potentially be disturbing or hazardous to others.) 
\newline
Note: The scene description does not explicitly mention the scooter being used as a weapon, theft, or skateboarders colliding, but riding a scooter on a crosswalk raises concerns about predictable and safe movement, and potentially disturbing public peace.
}
\end{appendices}

\end{document}